\documentclass{article}
\usepackage{spconf,amsmath,graphicx}

\usepackage{graphics,cite,graphicx,amsfonts,flafter}
\usepackage{amsmath,amssymb,url,verbatim,graphicx}
\usepackage{subfig,float,cases,mathptmx,epsfig,epsf,wrapfig}
\usepackage{xfrac,caption,array,supertabular}
\usepackage{textcomp,multirow,multicol}
\usepackage{latexsym,siunitx,booktabs}
\usepackage[section]{placeins}
\usepackage[linesnumbered,ruled,vlined]{algorithm2e}
\usepackage{xcolor}
\usepackage{bbm}
\def\be{\begin{eqnarray}}
\def\ee{\end{eqnarray}}
\def\ben{\begin{eqnarray*}}
\def\een{\end{eqnarray*}}

\def\bc{\begin{center}}
\def\ec{\end{center}}

\newcommand{\eqq}{$\,=\,$}
 
\newcommand{\bx}{{\textbf{x}} } 
\newcommand{\by}{{\textbf{y}} } 
 
\usepackage{titlesec}
\usepackage[noend]{algpseudocode}
\usepackage[linesnumbered,ruled,vlined]{algorithm2e}
\usepackage{mathtools, cuted}

\SetCommentSty{mycommfont}
\usepackage{multicol}
\usepackage{lipsum}
\SetKwInput{KwInput}{Input}                
\SetKwInput{KwOutput}{Output}              

\usepackage{titlesec}
\titlespacing*{\section}{0pt}{0.4\baselineskip}{\baselineskip}

\title{A COMPLETE RECIPE FOR BAYESIAN KNOWLEDGE TRANSFER: OBJECT TRACKING} 
\name{Bahman Moraffah and Antonia Papandreou-Suppappola\thanks{Thanks to XYZ agency for funding.}}
\address{
{\small{School of Electrical, Computer, and Energy Engineering, Arizona State University, Tempe Arizona 85281}}\\ {\small{Emails: bahman.moraffah@asu.edu, papandreou@asu.edu}}}
\begin{document}

%
\maketitle
\begin{abstract}
The problem of sequentially transferring from a source object track and a model to another Bayesian filter has become ubiquitous. Due to the lack of a structural model that can capture the dependence among different models, the transfer may not be fully specified. In this paper, we introduce a novel Bayesian model that accounts for the model-jump from which the object can choose a model and follow.  We aim to track the trajectory of the object while sequentially transferring from the source object to the target object.  The main idea is to impute the dynamical model while tracking the object and estimating the state parameters of the moving object according to discretized dynamic systems. We demonstrate this procedure can handle the model mismatch as it sequentially corrects the predictive model.  Particularly, for a fixed number of motion models, the object can learn what motion to follow at each time step. We employ a prior model for each model and then adaptively correct for changing one model to another to robustly estimate object trajectory under various motions.  More concretely, we propose a robust Bayesian recipe to handle the model-jump and then integrate it with a Markov chain Monte Carlo (MCMC) approach to sample from the posterior distribution. We demonstrate through experiments the advantage of accounting for model-jump in our proposed method for knowledge transfer between learning tasks in Bayesian transfer learning.

\end{abstract}
\section{{\large{Introduction}}}
\label{sec:intro}

Tracking performance can highly be affected by unknown variations in environmental 
conditions\cite{bar1990tracking}. Transfer learning is one of the fundamental concepts that can alleviate this issue by learning and transferring information  
between tracking sources \cite{Pan10,Tor10}. 
In \cite{Paz18},  authors improved Bayesian transfer learning by sequentially transferring 
  an optimal joint distribution model of the states and measurements. Pape\v{z} et al. then proposed a probabilistic technique that relaxes the target observation model using a scale-mixing parameter for transferring the first and second moments of the source data predictor \cite{papevz2020bayesian}.  A three-step extension of the Kalman filter was introduced to address issues in Bayesian knowledge transfer from a secondary to a primary filter \cite{foley2018}. Transferring knowledge between Kalman filters with the use of local variational Bayes approximation was also introduced in \cite{milan2018, milan2019}. This technique introduces a positive transfer of the source knowledge, however, it suffers from the lack of model flexibility.  In \cite{Jai17},  instead of learning a single dynamic model, models 
are learned at each source using online Bayesian moment matching. Recently,  we separately learned and transferred the parameters of 
  probabilistic models to account for unknown and time-varying measurement noise 
  conditions at the primary source  \cite{Alotaibi2020}.
These frameworks, however, lack to capture full model dependency due to the model-jumps from one time-step to another. And thus, they may not fully transfer the knowledge which leads to lower accuracy and model mismatch in state estimation. 

On the other hand, on account of advances in Markov chain Monte Carlo (MCMC) techniques for high dimensional data, Bayesian techniques in object tracking have become ubiquitous \cite{moraffah2019, moraffah2018 }. In this paper, we particularly propose a fully Bayesian knowledge transfer technique to account for a model correction method for which several varying dynamical systems can be accurately chosen.  In particular, our method provides a full recipe for tracking a moving object under time-varying dynamics by incorporating the model updates into the dynamical systems. The main advantage of this model over existing methods lies in the fact that model-jump is learned through the process and we no longer need to be concerned about the possible behaviors of objects over time under numerous conditions.  Moreover, the generality of this technique - any dynamic system can be accounted for-- provides a robust fully Bayesian technique that can successfully transfer knowledge between various dynamic systems. This framework is flexible and can be incorporated into a multi-object tracking setting. 
 
The rest of the paper is organized as follows. Section \ref{sec:PF} describes problem setting and imposed assumptions. Section \ref{sec:PM} provides detailed information on our proposed Bayesian knowledge transfer algorithm to track an object in the presence of multiple models. In Section \ref{sec:sim}, the performance of the algorithm as well as the use of this algorithm in a nonlinear setting is demonstrated through experiments.
 
 \section{{\large{Problem Formulation }}}
\label{sec:PF}
  Consider the problem of learning of source while we learn the discretized dynamical model to track a moving object. The goal is to estimate the unknown state parameters of the objects according to the corresponding time-varying dynamic system. To this end, we assume that object can only choose from a finite number of dynamical systems at each time, and continue its motion i.e., there are $L$ models at time $k$ ($L$ equations of motions at time $k$) $\mathcal{M}_k = \{\mathcal{M}_{j,k}, j = 1, 2, \dots, L\}$.  Let the initial probability of selecting each model be uniform, that is, $\mathbb{P}(\mathcal{M}_{j,0})= 1/L$ for all $j$. It is worth mentioning, these assumptions are not restrictive as the object can freely explore the scene and hence there is no advantage for one model over another.


For the sake of brevity, we assume there is no birth and death of objects. However, this method can simply be generalized to the case of the object leaving the field of view and new objects coming into the scene. One can simply integrate multiple object modeling for tracking introduced in \cite{moraffah2019PY, moraffah2018, moraffah2019} into our proposed method to manage model-jumps for each object. Since the model-jump is imputed through the process, this modeling elucidates the model mismatch.

 \section{{\large{Bayesian Knowledge Transfer Algorithm}}}
\label{sec:PM}

We propose a novel Bayesian method that can adaptively switch among models at each time to account for the different behavior of the object. This model fully captures the dependence among models and learns the appropriate model through the inference process. In addition to model mismatch justification,  this algorithm can accurately and robustly estimate the object trajectory in highly non-linear models. Moreover, this algorithm is flexible as it can simply be generalized to multiple objects with an unknown time-varying number of objects.  In what follows, we provide the details for the proposed algorithm. We first describe how models transition and then given the generated measurements, we impute the model and estimate the target trajectory. 

\subsection{{\normalsize Transition Model}}
Let $x_{j,k}$ be the state of the object at time $k$ following model $j$ which is transitioned from model  $j'$ according to the state transition equation
\begin{flalign}
\label{eq:motion}
x_{j,k} = g_{j',j}(x_{j', k-1}, \mathcal{M}_{j', k-1}) + v_{j'\to j, k}
\end{flalign}
where $v_{k,j'\to j}$ are independent noise that depends on the transformation from model $j'$ to model $j$. The function $g_{j',j}$ is the transition function that in practice often comes from the physical model defining each dynamic system. Define $\{\theta_{j,k}\} \in \Theta_{j,k}\subset\mathbb{R}^{N_x}$ to be the set of all parameters associated with model $j$ at time $k$. The equation (\ref{eq:motion}) can then equivalently be written as 
\begin{flalign}
 p(x_{j,k}, \mathcal{M}_{j,k} \mid \Theta_{j', k-1}, x_{j', k-1}, \mathcal{M}_{j', k-1})
 \end{flalign}
for some distribution $P$ with density $p$. 

Note that this model assumes that the transition between models is known in advance. In practice, this is however not the case and transitions are not known a priori.  To address this issue, we define a diffeomorphism (differentiable and invertible map) that can correct for model-jumps. Let $h_{j',j}: \{1,\dots, L\}\times \Theta_{j'} \to \{1,\dots, L\}\times \Theta_{j} $ be a diffeomorphism map between space of parameters of $\mathcal{M}_{j'}$ and $\mathcal{M}_{j}$ such that
\begin{flalign}
(j, \theta_j) = h_{j',j}(j',\theta_{j'}).
\end{flalign}
 We assume homogeneity in model transition, meaning the transition between models is time-independent.
It is worth noting that for this model to be well-defined $h_{j',j}$ needs to be bijective (invertible) with respect to the first argument so that one can go backward and forward between models. We require function $h$ to be differentiable for MCMC sampling purposes. Note that $h$ can be any function that satisfies the aforementioned conditions. For instance, $h_{j',j}$ can be a mapping that assigns any $j'$ to any $j$ with initial probability $J(j'\to j) = 1/L$ and draws $\theta_j$ according to the Kalman Filter, i.e., $\theta_j | \theta_{j'} \sim \mu(\theta_{j'}, \theta_j)$, where $\mu$ is Gaussian.

\subsection{{\normalsize Measurement Model}}
 Assume that measurements at time $k$ are generated according to the measurement equation at time $k$ that follows  
\begin{flalign}
\label{eq:measurement}
y_{j, k} = T(x_{j,k}) + w_{j, k} 
\end{flalign}
where $y_{j,k}, w_{j,k}$ are the measurements and noise given model $j$, respectively. Similarly, let $\{\phi_{j,k}\} \in \Phi_{j,k}\subset \mathbb{R}^{N_y}$ be the parameters of the measurement given model $j$, we can similarly re-write equation (\ref{eq:measurement}) as 
\begin{flalign}
 f( y_{j, k} \mid \Phi_{j,k}, x_{j,k}, \mathcal{M}_{j,k})
\end{flalign}
for some distribution $F$ with density $f$. 

Let $y_{j,k} = \{y^1_{j,k}, \dots, y^{M_k}_{j, k}\}$ be the set of $M_k$ observations at time step $k$ given the model $\mathcal{M}_j$. For simplicity, we assume $f$ follows a Gaussian mixture model with parameters $\{\phi^m_{j,k}\} = \{\mu^m_{j,k}, \Sigma^m_{j,k}\}$ -- set of mean and covariance matrices. The Bayesian hierarchical model, given model $\mathcal{M}_j$, for our measurements at time $k$ is outlined as follows:
{{ \footnotesize
 \setlength{\arraycolsep}{2.2pt}
\medmuskip = 1mu 
\begin{flalign}
\label{eq:gmmmodel}
&\pi_{k,\ell} \sim\text{Dir}(1/C_k, \dots, 1/C_k), \hspace{2cm} \ell = 1,2,\dots, C_k\notag\\
&c_{\ell, k} \sim \text{Cat}(\pi_\ell)\\
&\phi^m_{j, k}\mid \mathcal{M}_{j,k} \sim {\mathrm  {NIW}}({\boldsymbol  m}_{j,k},\lambda_{j,k} ,{\boldsymbol  \Psi_{j,k} },\nu_{j,k} )\hspace{2cm} m = 1, \dots, M_k\notag\\
&y^m_{j,k}|\phi^m_{j,k}, x_{j,k}, \mathcal{M}_{j,k} \sim \text{GMM}(\{\pi_{\ell,k}\}, \{\phi^{c_{\ell,k}}_{j, k}\}\mid x_{j,k})\notag
\end{flalign}}}
\noindent where $\{{\boldsymbol  m}_{j,k},\lambda_{j,k} ,{\boldsymbol  \Psi_{j,k} },\nu_{j,k}\}$ and $C_k$ are fixed parameters of Normal-inverse-Wishart (NIW) for model $j$ and  the number of clusters in GMM at time $k$, respectively. In case the number of clusters in GMM is not known {\it a priori}, we can replace this step with the Dirichlet process mixture model (DPMM) to adjust for the unknown number of clusters. In our experience, choosing $C_k$ according to the Bayesian information criterion (BIC) score and DPMM provides almost identical results in this problem. 
 
\section{{\large{Inference}}}
In this section, we aim to estimate the parameters, latent variables, models,  as well as object trajectory, i.e., the state variables at each time $k$, conditioned on the observations up to time $k$. It is worth mentioning we are only interested in the state estimation and models and the rest of the parameters and latent variables are nuisances. Nonetheless, for the sake of completeness, we provide the full posterior distribution. One can simply marginalize the unwanted variables given the representation of the model, the graphical model representing this model. The posterior distribution $p(X, \Theta, \Phi , \mathcal{M}\mid Y)$, where $X = \{{x^u_{j,k}}\}_{u, j, k}$, $\Theta = \{\Theta_{j,k}\}_{j,k}$, $\Phi = \{\Phi_{j,k}\}_{j,k}$, $\mathcal{M} =  \{\mathcal{M}_{k}\}_k$, and $Y = \{Y_{j,k}\}_{j,k}$ can be expanded as follows:
{\small
\be
\begin{split}
\label{eq:posterior}
&p(X, \Theta, \Phi , \mathcal{M}\mid Y)  \propto  p(X, \Theta, \Phi , \mathcal{M}, Y) \\
& = p(\mathcal{M}) p(\Theta\mid \mathcal{M}) p(\Phi\mid \mathcal{M}) p(X|\mathcal{M},\Theta) p(Y\mid X, \mathcal{M}, \Phi) \\
& = p(\mathcal{M}) \prod_{k=1}^K \prod_{j, j' = 1}^L \Bigg\{\\
&\resizebox{195pt}{!}{$\Big[\mu(\theta_{j', k-1}, \text{Proj}_2(h_{j',j}(j',\theta_{j',k-1})))\Big]^{\mathbbm{1}(\mathcal{M}_k = \text{Proj}_1(h_{j',j}(j',\theta_{j',k-1}),\mathcal{M}_{k-1} = j') }$}\\
&\prod_{u = 1}^{U_k} \big[p(x^u_{j,k}|x_{j',k-1}, \theta^j_{k})\big]^{\mathbbm{1}(\mathcal{M}_k = \text{Proj}_1(h_{j',j}(j',\theta_{j',k-1}),\mathcal{M}^{k-1} = j') }\\
& \prod_{m=1}^{M_k}\big[p(\phi^m_{j,k})p(y^m_{j,k}|\phi^m_{j,k}, x_{j,k})\big]^{{\mathbbm{1}(\mathcal{M}_k = \text{Proj}_1(h_{j',j}(j',\theta_{j',k-1}))}}\Bigg\}
\end{split}
\ee}
\noindent where $\text{Proj}_i$ is the projection map onto the $i$th component, $i = 1,2$, $\mathbbm{1}(\cdot)$ is an indicator function; i.e., the indicator function is $1$ when the argument is true and otherwise it is zero, and  $U_k$ is the cardinality of state space which may vary from one time to another as dynamic systems change over time. Note that all terms in the posterior equation (\ref{eq:posterior}) are defined through equations (\ref{eq:motion}) - (\ref{eq:gmmmodel}). Computing the exact posterior mean is impossible as there is no closed-form solution for the integral. To be able to do inference and estimate the object trajectory, we take advantage of the Markov chain Monte Carlo methods. To perform fast and accurate sampling from the posterior distribution (\ref{eq:posterior}), we combine Monte Carlo methods. To facilitate quicker mixing through the joint distribution, we exploit Hamiltonian Monte Carlo (HMC) to sample from continuous variables and utilize Gibbs sampling for discrete variables. This framework is outlined in Algorithm \ref{alg:alg1}.

\section{{\large{Experimental Results}}}
 \begin{algorithm}[t!]
\DontPrintSemicolon
  \KwInput{ $\mathcal{M}_{j',k-1} $, $\mu(\theta_{j',k-1},\cdot)$, $h_{j',j}$, \text{configurations at time} $k-1$ }
  \KwOutput{$\hat{\mathcal{M}}_{j,k}, \hat{x}_{j,k}$}
  \KwData{$\{Y_{j,k}\}_{j,k}$}
  
 \For {j, j' = 1, \dots, L}
 {
 \For {$m = 1,\dots, M_k$}
 { Compute the likelihood according to \ref{eq:gmmmodel}\;
  Compute posterior distribution Equation \ref{eq:posterior} treating $x_{j,k}$ and $\mathcal{M}_{j,k}$ as latent variables\;
  Sample from Equation \ref{eq:posterior} using HMC and Gibbs sampling\;
  Compute posterior predictive distribution given samples obtained from HMC and Gibbs sampler}

}
Estimate $\hat{\mathcal{M}}_{j,k}$\;
Estimate $\hat{x}_{j,k} =  \mathbb{E}_{x_k|Y, \hat{\mathcal{M}}_{j,k}}[x_{j,k}]$

\caption{{\small{Bayesian Knowledge Transfer Algorithm}}}
\label{alg:alg1}
\end{algorithm}
In this section, we provide two sets of experiments to demonstrate the advantage of our proposed Bayesian knowledge transfer algorithm over existing methods. Throughout this section, the measure of accuracy for object trajectory estimation is the mean squared error (MSE) defined as
 \be
 \text{MSE} = \frac{1}{N} \sum_{i = 1}^{N}||x^{\text{true}}_i - \mathbb{E}_{x_i|Y, \mathcal{M}}[x_i]||^2
 \ee
where $\mathbb{E}_{x_i|Y, \mathcal{M}}[x_i]$ is the expected value taken with respect to the posterior distribution given model $\mathcal{M}$ and $||\cdot||$ is the Euclidean norm. We utilized HMC and Gibbs sampler to sample from the posterior distribution provided in (\ref{eq:posterior}).  We run 10,000 Monte Carlo realizations.
\label{sec:sim}
\subsection{{\normalsize Experiment I: Evaluation}}
To provide a measure of performance, we assume the following linear state-state model where the object can uniformly at random select among $M = 3$ models:
\begin{flalign}
\begin{split}
\bx_{j, k} &= A x_{j', k-1} + v_{j'\to j, k} \hspace{2cm} j, j' \in \{1,2,3\}\\
\by^j_k &= C x_{j, k} + w_{j,k}
\end{split}
\end{flalign}
where  $v_{j'\to j, k} \sim \mathcal{N}(0, Q_j)$ and $w_{j,k} \sim \mathcal{N}(0, R_j)$ are noise variables associated with the state variables and the observations such that 
\begin{flalign}
\begin{split}
 A= \begin{bmatrix}
1 & \Delta \\
0 & 1 
\end{bmatrix},  C= \begin{bmatrix}
\mathbb{I}_2& 0_{2\times2} 
\end{bmatrix}\\
Q_j= \alpha \begin{bmatrix}
 j \Delta^2 /4& 0 \\
0 &j^2 \Delta/ 3 
\end{bmatrix}, R_j= j \beta \mathbb{I}_2
\end{split}
\end{flalign}
 \noindent where $\Delta = 0.1$, $\alpha = 0.01$, $\beta = 0.5$, and $\mathbb{I}$ is the identity matrix. 
 
   \begin{figure}[th!]
    \centering
    \includegraphics[width=0.49\textwidth]{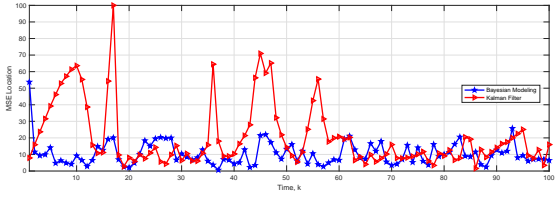}
    \caption{MSE comparison for the location estimation. MSE for our Bayesian techniques (blue) is compared to MSE for the Kalman filter (red). }
    \label{fig:mse}
\end{figure}
 
\begin{figure}[th!]
    \centering
    \includegraphics[width=0.49\textwidth]{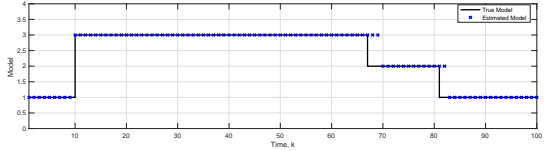}
    \caption{Model selection using Bayesian knowledge transfer algorithm vs true model. }
    \label{fig:modelselec}
\end{figure}

 Let $\bx_{j, k} = (x_{j, k}, y_{j, k}, \dot{x}_{j, k}, \dot{y}_{j, k})$ be the state parameters where $(x_{j, k}, y_{j, k})$
 and $(\dot{x}_{j, k}, \dot{y}_{j, k})$ are the position and velocity, respectively. Computing the mean squared error (MSE) of our proposed technique and comparing it to that of the proposed technique in \cite{milan2019}, we demonstrate the advantage of our technique over the Kalman filter techniques introduced in \cite{milan2018, milan2019}, Figure \ref{fig:mse}.   For the sake of simplicity, we assumed the prior on all measurement parameters are time-independent. In particular, we use $\text{NIW}(0.001j, 0, 100, j \mathbb{I})$ for models $j = 1, 2, 3$ and utilize BIC to choose $C_k$ at each time step $k$.  It is worth mentioning that our proposed Bayesian method not only outperforms the Kalman filter-based techniques but it also has superior performance for nonlinear dynamical systems which will be demonstrated in the full paper. Figure \ref{fig:modelselec} depicts the selected model through our Bayesian algorithm against the true model. 

\subsection{{\normalsize {Experiment II: System with Turn}} }

In this section, we turn our attention to a complicated situation where the object can follow one of the ten models where each model may not follow a linear trajectory. In particular, the object is moving in the two-dimensional (2-D) plane following the coordinated turn model. The state parameter vector is 
\ben
\bx_{j, k} =  [x_{j, k} \   \dot{x}_{j, k}\   y_{j , k}  \  \dot{y}_{j, k} \  \omega_{j, k}]^T, \hspace{0.2cm} j = 1, \dots, 10
\een
where $(x_{j, k},  y_{j, k})$ 
and $(\dot{x}_{j, k}, \dot{y}_{j , k})$  are the 2-D Cartesian coordinates for position and velocity, respectively, and  $\omega_{j, k}$ is the constant turn rate given model $j$. 
 
  Let the transition model be  $\bx_{j, k} =  F_j \, \bx_{j, k-1} +  v_{j'\to j, k}$ with $v_{j'\to j, k}$ being time-independent zero-mean Gaussian with  covariance matrix $Q_{j'\to j}$ such that
 {\footnotesize
 \setlength{\arraycolsep}{2.pt}
\medmuskip = 1mu 
 \ben
 F_j =  \begin{bmatrix}
1 &  A & 0 
& - B & 0\\
0  & C& 0 & -D & 0\\
0 & B
& 1 & A & 0\\
0 & D& 0 &C&0  \\
0 & 0 & 0& 0& 1 \end{bmatrix},  \ \ \
Q_{j'\to j} = \begin{bmatrix}
  \frac{\sigma^2}{4}  &  \frac{\sigma^2}{2} & 0 & 0 & 0 \\
 \frac{\sigma^2}{2} & {\scriptstyle \sigma^2} & 0 & 0 & 0 \\
 0 & 0 &  \frac{\sigma^2}{4}  &  \frac{\sigma^2}{2} & 0 \\
 0 & 0 &  \frac{\sigma^2}{2} & {\scriptstyle \sigma^2} & 0 \\
 0 & 0 &  0 & 0 & {\scriptstyle  \sigma_v^2} 
 \end{bmatrix}
\een}
{\small{$A = \sin(|j'-j|\omega_{k-1})/|j'-j|\omega_{k-1}$, $B = 1-\cos(|j'-j|\omega_{k-1})/|j'-j|\omega_{k-1}$, $C =  \cos(|j'-j|\omega_{k-1})$, $D = \sin(|j'-j|\omega_{k-1})$, $\sigma \eqq 15\log(|j-j'|)$ m/s$^2$}},  and {\small {$\sigma_v \eqq \sqrt{j}\pi/180$}} radians/s.

Given model $\mathcal{M}_j$, the measurements are {\small{$y_{j,k} \eqq [\phi_{j,k} \  r_{j,k}]^T  + w_{j,k}$}}, where $[\phi_{j,k} \  r_{j,k}] \eqq [ \arctan{\!(y_{j, k}/x_{j, k})} \ (x^2_{j, k} +  y^2_{j, k})^{1/2} ]$,   $\phi_{j,k} \!\!\in  \!\!(-\pi/2,  \pi/2)$  is the 
 bearing and  $r_{j,k}  \!\!\in  \!\!(0, 2)$ km is the range. Let $w_{j,k}$ be zero-mean 
 Gaussian with covariance matrix $Q_w\eqq \text{diag}(25, (\pi/180)^2)$, independent of the models.

   \begin{figure}[th!]
    \centering
    \includegraphics[width=8.5cm, height=3.3cm]{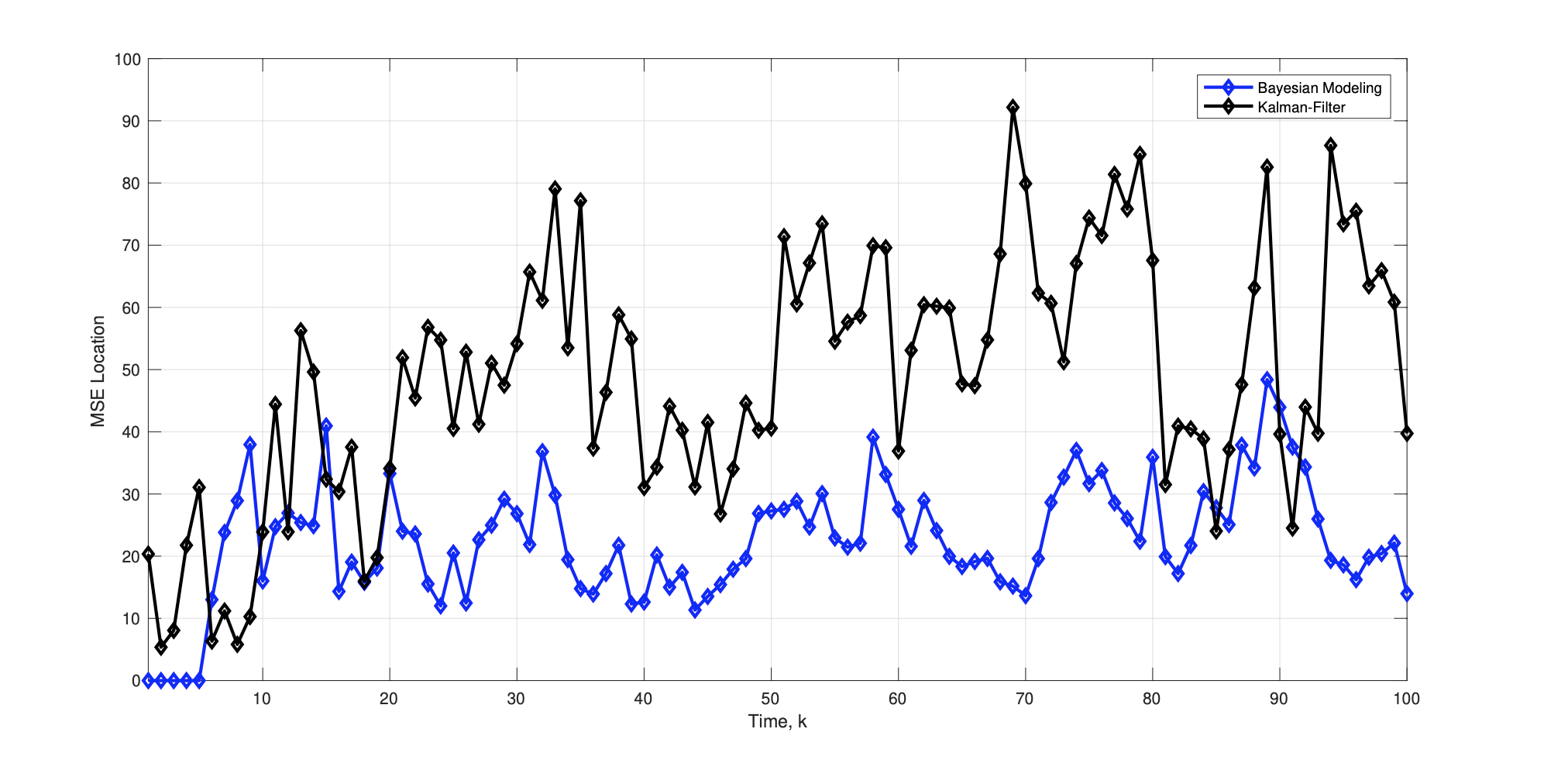}
    \caption{Mean squared error comparison between our proposed Bayesian approach and the Kalman Filter-based approach for object trajectory for 10 different models. }
    \label{fig:traj_complicated}
\end{figure}
 
\begin{figure}[th!]
    \centering
    \includegraphics[width=0.49\textwidth]{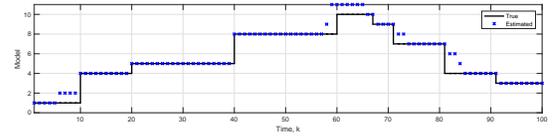}
    \caption{Model estimation for 10 different models. }
    \label{fig: model_2}
\end{figure}

 Figure \ref{fig:traj_complicated} displays the mean squared error for the above model using our proposed Bayesian knowledge transfer framework. This graph shows the performance of our proposed Bayesian method for a complicated system with a turn. Moreover, we estimated the model during the trajectory estimation process, Figure \ref{fig: model_2}. This graph demonstrates the accuracy of this technique even with an increasing number of models.
 
\section{{\large{Conclusion}}}

In this paper, we presented a full Bayesian knowledge transfer recipe to address the model-jumps for an object. At each step, the object can select a model, given diverse conditions, and follow it. Our Bayesian model successfully accounts for the model mismatch as it learns the best model at each time step. We demonstrated through simulations that our proposed Bayesian knowledge transfer algorithm is efficient and can accurately estimate the model and therefore estimate the trajectory of the object in the presence of sudden model change. Due to the flexibility and efficiency of this method, one can simply extend this framework to a multi-object model that follows a birth and death process.

\bibliographystyle{IEEEbib}

\bibliography{ref_BKT}

\begin{thebibliography}{10}

\bibitem{bar1990tracking}
Yaakov Bar-Shalom, Thomas~E Fortmann, and Peter~G Cable,
\newblock ``Tracking and data association,'' 1990.

\bibitem{Pan10}
S.~J. Pan and Q.~Yang,
\newblock ``A survey on transfer learning,''
\newblock {\em IEEE Transactions on Knowledge and Data Engineering}, vol. 22,
  pp. 1345--1359, 2010.

\bibitem{Tor10}
L.~Torrey and J.~Shavlik,
\newblock ``Transfer learning,''
\newblock in {\em Handbook of Research on Machine Learning Applications and
  Trends: Algorithms, Methods, and Techniques}, E.~S. Olivas, J.~D.~M.
  Guerrero, M.~M. Sober, J.~R.~M. Benedito, and A.~J.~S. Lopez, Eds.,
  chapter~11, pp. 242--264. Information Science Reference, 2010.

\bibitem{Paz18}
Milan Pape{\v{z}} and Anthony Quinn,
\newblock ``Dynamic {B}ayesian knowledge transfer between a pair of {K}alman
  filters,''
\newblock in {\em IEEE International Workshop on Machine Learning for Signal
  Processing}, 2018, pp. 1--6.

\bibitem{papevz2020bayesian}
Milan Pape{\v{z}} and Anthony Quinn,
\newblock ``Bayesian transfer learning between student-t filters,''
\newblock {\em Signal Processing}, vol. 175, pp. 107624, 2020.

\bibitem{foley2018}
Conor Foley and Anthony Quinn,
\newblock ``Fully probabilistic design for knowledge transfer in a pair of
  kalman filters,''
\newblock {\em IEEE Signal Processing Letters}, vol. 25, no. 4, pp. 487--490,
  2018.

\bibitem{milan2018}
Milan Papež and Anthony Quinn,
\newblock ``Dynamic bayesian knowledge transfer between a pair of kalman
  filters,''
\newblock in {\em 2018 IEEE 28th International Workshop on Machine Learning for
  Signal Processing (MLSP)}, 2018, pp. 1--6.

\bibitem{milan2019}
Milan Pape{\v{z}} and Anthony Quinn,
\newblock ``Robust bayesian transfer learning between kalman filters,''
\newblock in {\em 2019 IEEE 29th International Workshop on Machine Learning for
  Signal Processing (MLSP)}. IEEE, 2019, pp. 1--6.

\bibitem{Jai17}
P.~Jaini, Z.~Chen, P.~Carbajal, E.~Law, L.~Middleton, K.~Regan, et~al.,
\newblock ``Online {B}ayesian transfer learning for sequential data modeling,''
\newblock in {\em International Conference on Learning Representions}, 2017.

\bibitem{Alotaibi2020}
Omar Alotaibi and Antonia Papandreou-Suppappola,
\newblock ``Transfer learning with bayesian filtering for object tracking under
  varying conditions,''
\newblock in {\em 2020 54th Asilomar Conference on Signals, Systems, and
  Computers}, 2020, pp. 1523--1527.

\bibitem{moraffah2019}
B.~Moraffah and A.~Papandreou-Suppappola,
\newblock ``Random infinite tree and dependent {P}oisson diffusion process for
  nonparametric {B}ayesian modeling in multiple object tracking,''
\newblock in {\em International Conference on Acoustics, Speech, and Signal
  Processing}, 2019, pp. 5217--5221.

\bibitem{moraffah2018}
B.~Moraffah and A.~Papandreou-Suppappola,
\newblock ``Dependent {D}irichlet process modeling and identity learning for
  multiple object tracking,''
\newblock in {\em Asilomar Conference on Signals, Systems, and Computers},
  2018, pp. 1762--1766.

\bibitem{moraffah2019PY}
Bahman Moraffah, Antonia Papandreou-Suppappola, and Muralidhar Rangaswamy,
\newblock ``Nonparametric {B}ayesian methods and the dependent {P}itman-{Y}or
  process for modeling evolution in multiple object tracking,''
\newblock in {\em 2019 22th International Conference on Information Fusion
  (FUSION)}. IEEE, 2019, pp. 1--6.

\end{thebibliography}

\end{document}